\def\year{2020}\relax
\documentclass[letterpaper]{article} 
\usepackage{aaai20}  
\usepackage{times}  
\usepackage{helvet} 
\usepackage{courier}  
\usepackage[hyphens]{url}  
\usepackage{graphicx} 
\usepackage{url}             
\usepackage{booktabs}        
\usepackage{amsfonts}        
\usepackage{nicefrac}        
\usepackage{microtype}       
\usepackage{graphicx}        
\usepackage{subfigure}
\usepackage{enumitem}
\usepackage{multirow}

\usepackage{algorithm}
\usepackage{algorithmic}

\usepackage{amsmath}
\usepackage{amsthm}
\usepackage{amssymb}
\usepackage{color}
\usepackage{float}
\usepackage{lipsum}
\newtheorem{theorem}{Theorem}

\newtheorem{proposition}[theorem]{Proposition}

\newtheorem{remark}{Remark}

\newsavebox{\measurebox}

\usepackage{multirow}
\usepackage{array}
\newcolumntype{L}[1]{>{\raggedright\let\newline\\\arraybackslash\hspace{0pt}}m{#1}}
\newcolumntype{C}[1]{>{\centering\let\newline  \\\arraybackslash\hspace{0pt}}m{#1}}
\newcolumntype{R}[1]{>{\raggedleft\let\newline \\\arraybackslash\hspace{0pt}}m{#1}}

\newcommand{\Px}[2]{\text{\normalfont prox}_{#1}( #2 ) }
\newcommand{\NM}[2]{\left\|  #1 \right\|_{#2}}
\title{Efficient Neural Architecture Search 
	via Proximal Iterations }
\author{
	Quanming Yao\textsuperscript{\rm 1}\thanks{Q. Yao and J. Xu contribute equally,
    and the work was performed when J. Xu was an intern in 4Paradigm.},
	Ju Xu\textsuperscript{\rm 3}\footnotemark[1],
	Wei-Wei Tu\textsuperscript{\rm 1},
	Zhanxing Zhu\textsuperscript{\rm 2,3,4}\thanks{Zhanxing Zhu is the corresponding author.}\\
	\textsuperscript{\rm 1}4Paradigm Inc,
    \textsuperscript{\rm 2}School of Mathematical Sciences, Peking University\\
    \textsuperscript{\rm 3}Center for Data Science, Peking University,
    \textsuperscript{\rm 4}Beijing Institute of Big Data Research (BIBDR)\\
     \{yaoquanming, tuweiwei\}@4paradigm.com, \{xuju, zhanxing.zhu\}@pku.edu.cn
}

\begin{document}
\maketitle
\begin{abstract}
Neural architecture search (NAS) 
attracts much research attention because of its ability to identify better architectures than handcrafted ones. 
Recently, 
differentiable search methods become the state-of-the-arts on NAS, 
which can obtain high-performance architectures in several days. 
However, 
they still suffer from huge computation costs 
and inferior performance due to the construction of the supernet.
In this paper, 
we propose an efficient NAS method based on proximal iterations (denoted as NASP). 
Different from previous works, 
NASP reformulates the search process as an optimization problem with a discrete constraint on architectures
and a regularizer on model complexity.
As the new objective is hard to solve, 
we further propose an efficient algorithm inspired by proximal iterations for optimization.
In this way,
NASP is not only much faster than existing differentiable search methods,
but also can find better architectures and balance the model complexity.
Finally,
extensive experiments on various tasks demonstrate that NASP can obtain high-performance architectures 
with more than 10 times speedup over the state-of-the-arts.
\end{abstract}

\section{Introduction}
\label{sec:intro}

\begin{table*}[t]
	\centering
	\caption{Comparison of the proposed NASP with other state-of-the-art NAS methods
		on four perspectives of searching:
		differentiable (denoted as ``diff''),
		cell,
		complete,
		and constraint.}
	\small
	\begin{tabular}{ c|C{30px}|C{30px}|c|c||c||c}
		\hline
		                                     & \multicolumn{2}{c|}{space} & \multirow{2}{*}{complete} & complexity &        discrete        & \multirow{2}{*}{search algorithm} \\
		                                     &   diff   &      cell       &                           &  control   &     architectures      &                                   \\ \hline
		NASNet-A \cite{zoph2017learning}     & $\times$ &     $\surd$     &          $\surd$          &  $\surd$   &        $\times$        &      reinforcement learning       \\ \hline
		AmoebaNet \cite{real2018regularized} & $\times$ &     $\surd$     &          $\surd$          &  $\surd$   &        $\times$        &        evolution algorithm        \\ \hline
		SNAS \cite{xie2018snas}              & $\surd$  &     $\surd$     &         $\times$          &  $\surd$   &        $\surd$         &      reinforcement learning       \\ \hline
		DARTS (Liu, et.al. 2019)             & $\surd$  &     $\surd$     &         $\times$          &  $\times$  &        $\times$        &         gradient descent          \\ \hline\hline
		NASP (proposed)                      & $\surd$  &     $\surd$     &          $\surd$          &  $\surd$   &        $\surd$         &        proximal algorithm         \\ \hline
	\end{tabular}
	\label{tab:nasalgs}
\end{table*}

Deep networks have been applied to many applications,
where proper architectures are extremely important to ensure good performance.
Recently,
the neural architecture search (NAS) 
\cite{zoph2017neural,baker2017designing}
has been developed as a promising approach to 
replace human experts on designing architectures,
which can find networks with fewer parameters and better performance
\cite{yao2018taking,automl_book}.
NASNet \cite{zoph2017neural} is the pioneered work along this direction and it models the design of convolutional neural networks (CNNs) as 
a multi-step decision problem and solves it with reinforcement learning \cite{sutton1998reinforcement}.

However,
since the search space is discrete and extremely large,
NASNet requires a month with hundreds of GPU to obtain a satisfying architecture.
Later,
observing the good transferability of networks
from small to large ones,
NASNet-A \cite{zoph2017learning} proposed to cut the networks into blocks
and then the search only needs to be carried within such a block or cell.
The identified cell is then used as a building block to assemble large networks.
Such two-stage search strategy dramatically reduces the size of the search space,
and subsequently leads to the significant speedup of various previous search algorithms
(e.g., evolution algorithm \cite{real2018regularized},
greedy search \cite{liu2018progressive},
and reinforcement learning \cite{zhong2018practical}).

Although the size of search space is reduced,
the search space is still discrete that is generally hard to be efficiently searched \cite{parikh2013proximal}.
More recent endeavors focused on how to change the landscape of the search space
from a discrete to a differentiable one \cite{luo2018neural,liu2018darts,xie2018snas}.
The benefit of such idea is that 
a differentiable space enables computation of gradient information,
which could speed up the convergence of underneath optimization algorithm.
Various techniques have been proposed,
e.g., DARTS \cite{liu2018darts} smooths design choices with softmax and trains an ensemble of networks;
SNAS \cite{xie2018snas} enhances reinforcement learning with a smooth sampling scheme.
NAO \cite{luo2018neural} maps the search space into a new differentiable space with an auto-encoder.

Among all these works (Tab.~\ref{tab:nasalgs}),
the state-of-the-art is DARTS \cite{liu2018darts}
as it combines the best of both worlds,
i.e., fast gradient descent (differentiable search space)
within a cell (small search space).
However,
its search efficiency and performance of identified architectures are still not satisfying enough.
As it maintains a supernet during the search,
from the computational perspective, all operations need to be forward and backward propagated during gradient descent while only one operation will be selected.
From the perspective of performance,
operations typically correlate with each other \cite{xie2018snas},
e.g., a 7x7's convolution filter can cover a 3x3 one as a special case.
When updating a network's weights,
the ensemble constructed by DARTS during the search may lead to inferior architecture being discovered.
 Moreover,
as mentioned in \cite{xie2018snas},
DARTS is not complete (Tab.~\ref{tab:nasalgs}),
i.e., the final structure needs to be re-identified after the search. 
This causes a bias between the searched architecture and the final architecture,
and might lead to a decay on the performance of the final architecture.

In this work,
we propose NAS with proximal iterations (NASP)
to improve the efficiency and performance of existing differentiable search methods.
We give a new formulation and optimization algorithm to NAS problem,
which allows searching in a  differentiable space while keeping discrete architectures.
In this way,
NASP removes the need of training a supernet,
then speeds up the search and leads to better architectures.
Our contributions are
\begin{itemize}[parsep=0pt,partopsep=0pt]
\item Except for the popularly discussed perspectives of NAS,
i.e., 
search space,
completeness,
and
model complexity,
we identify a new and important one,
i.e., \textit{constraint on architectures} (``discrete architectures'' in Tab.~\ref{tab:nasalgs}),
to NAS.

\item We formulate NAS as a constrained optimization problem,
which keeps the space differentiable but enforces architectures being discrete during the search,
i.e., only one of all possible operations to be actually employed during forward and backward propagation. 
This helps
improve searching efficiency
and decouple different operations during the training. 
A regularizer is also introduced into
the new objective,
which allows control of architectures' size.

\item Since such discrete constraint is hard to optimize
and simple adaptation of DARTS cannot be applied, 
we propose a new algorithm derived from the proximal iterations \cite{parikh2013proximal}
for optimization.
The closed-form solution to the proximal step with the proposed discrete constraint is new to
the optimization literature,
and removes the expensive \textit{second-order approximation} required by DARTS.
We further provide
a theoretical analysis to
guarantee convergence of the proposed algorithm. 

\item Finally,
experiments are performed with various benchmark data sets
on designing CNN and RNN architectures.
Compared with state-of-the-art methods,
the proposed NASP is not only much faster (more than ten times speedup over DARTS)
but also can discover better architectures.
These empirically demonstrate
NASP can obtain the state-of-the-art performance on both test accuracy and computation efficiency.
\end{itemize}
The implementation of NASP is available at:
\url{https://github.com/xujinfan/NASP-codes}.

\section{Related Works}

In the sequel,
vectors are denoted by lowercase boldface, 
and matrices by uppercase boldface.

\subsection{Proximal Algorithm (PA)}
\label{sec:relprox}

Proximal algorithm (PA) \cite{parikh2013proximal},
is a popular optimization technique in machine learning for handling constrained optimization problem as
$\min_{\mathbf{x}} f(\mathbf{x}), 
\;\text{s.t.}\;
\mathbf{x} \in \mathcal{S}$,
where $f$ is a smooth objective and $\mathcal{S}$ is a constraint set.
The crux of PA is the proximal step:
\begin{align}
\mathbf{x} = \Px{\mathcal{S}}{\mathbf{z}}
= \arg\min_{\mathbf{y}} \nicefrac{1}{2}\NM{\mathbf{y} - \mathbf{z}}{2}^2
\;\text{s.t.}\;
\mathbf{y} \in \mathcal{S}.
\label{eq:prox}
\end{align}
Closed-form solution for the PA update exists for many constraint sets in \eqref{eq:prox},
such as $\ell_1$- and $\ell_2$-norm ball \cite{parikh2013proximal}.
Then,
PA generates a sequence of $\mathbf{x}_t$ by
\begin{align}
\mathbf{x}_{t + 1} = \Px{\mathcal{S}}{\mathbf{x}_t - \varepsilon \nabla f(\mathbf{x}_t)},
\label{eq:proxiter}
\end{align}
where $ \varepsilon$ is the learning rate. PA guarantees to obtain the critical points of $f$ when 
$\mathcal{S}$ is a convex constraint,
and produces limit points when 
the proximal step can be exactly computed \cite{yao2017efficient}.
Due to its nice theoretical guarantee
and good empirical performance, 
it has been applied to many deep learning problems,
e.g., network binarization \cite{Bai2018}.

Another variant of PA with lazy proximal step \cite{xiao2010dual} maintains two copies of $\mathbf{x}$,
i.e.,
\begin{align}
\bar{\mathbf{x}}_t 
= \Px{\mathcal{S}}{\mathbf{x}_t },
\quad
\mathbf{x}_{t + 1} 
= \mathbf{x}_t - \varepsilon \nabla f(\bar{\mathbf{x}}_t),
\label{eq:proxiter:lazy}
\end{align}
which is also popularily used in deep learning for network quantization
\cite{courbariaux2015binaryconnect,hou2016loss}.
It does not have convergence guarantee in the nonconvex case,
but empirically performs well on network quantization tasks.
Finally,
neither \eqref{eq:proxiter} nor \eqref{eq:proxiter:lazy} have been introduced into NAS.

\subsection{Differentiable Architecture Search (DARTS)}
\label{sec:reldnas}

\begin{figure*}[ht]
	\centering
	\subfigure[Cell in NAS.]
	{\includegraphics[height=0.18\textwidth]{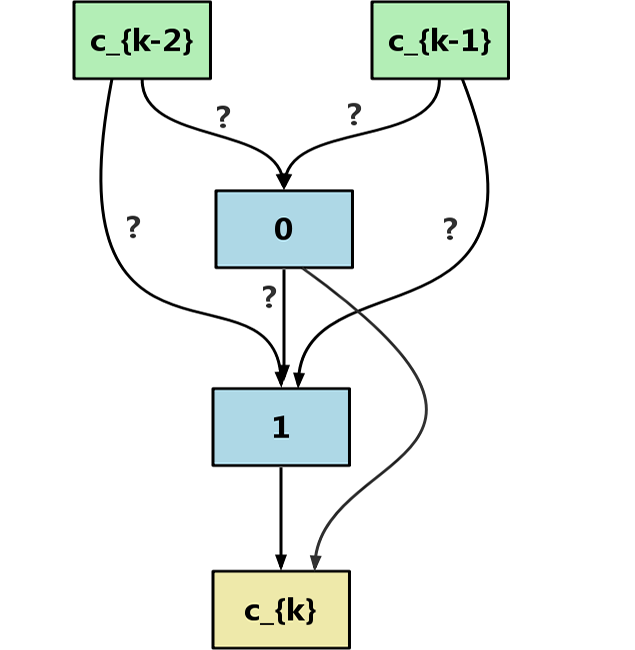}
		\label{fig:cell}}
	\qquad
	\subfigure[DARTS.]
	{\includegraphics[height=0.18\textwidth]{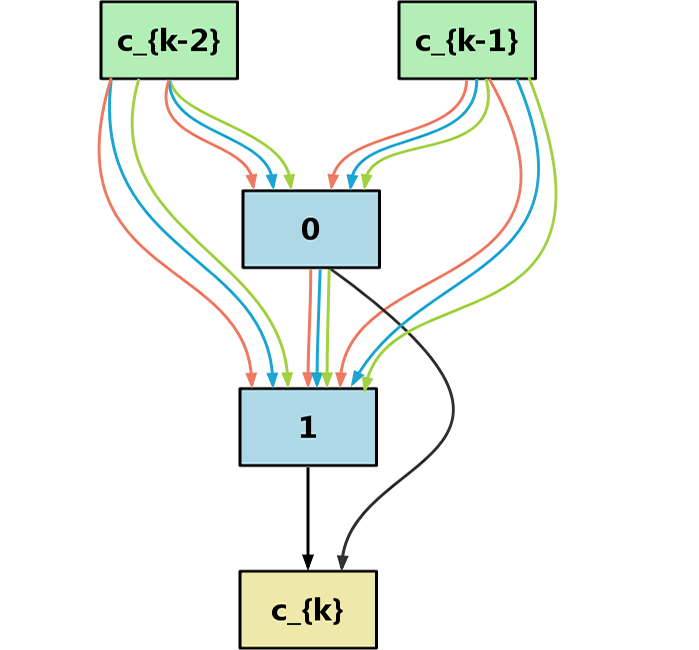}
		\label{fig:compdarts}}
	\qquad
	\subfigure[NASP.]
	{\includegraphics[height=0.18\textwidth]{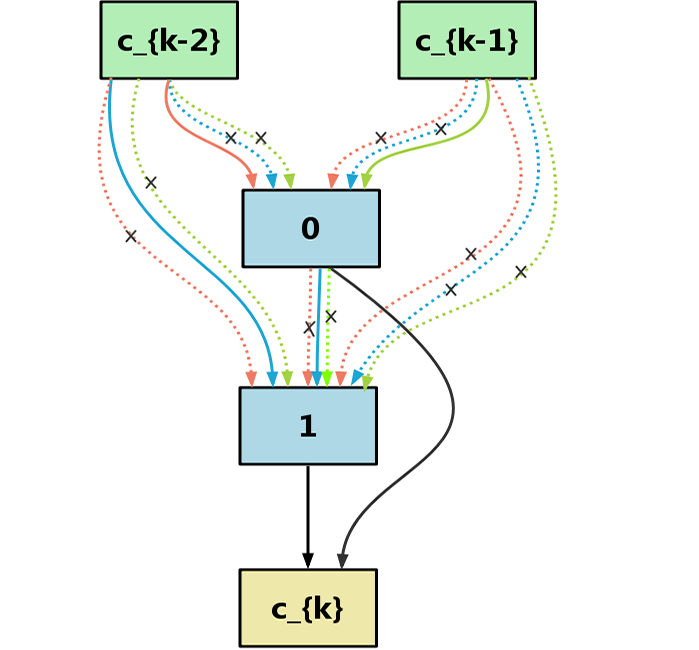}
		\label{fig:proxnas}}
	\qquad
	\subfigure[Parameter $\mathbf{A}$.]
	{\includegraphics[height=0.18\textwidth]{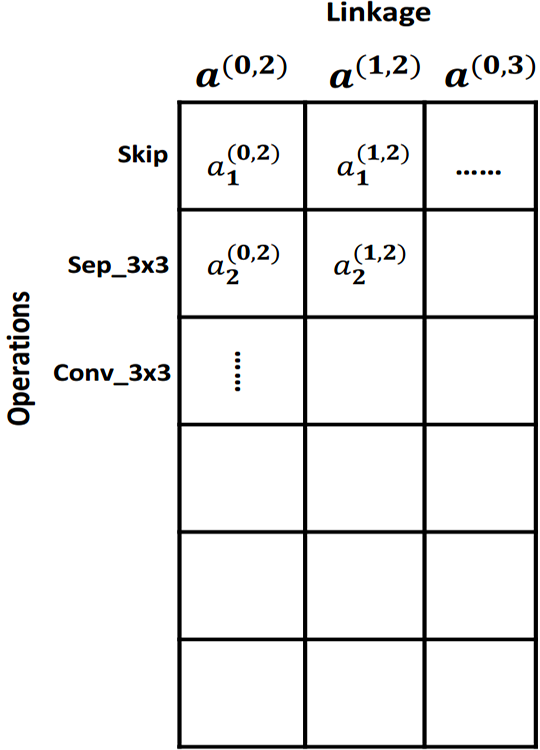}
		\label{fig:alpha}}
	
	\vspace{-10px}
	
	\caption{Comparison of computation graph in a cell between DARTS (Fig.~\ref{fig:compdarts}) and NASP (Fig.~\ref{fig:proxnas}).
		Three operations are considered,
		DARTS needs to forward and backward propagate along all operations for updating $w$,
		while NASP only requires computing along current selected one.
		The architecture parameters {\scriptsize $a^{(i,j)}_k$} can be arranged into a matrix form (Fig.~\ref{fig:alpha}).}
	\label{fig:darts}
\end{figure*}

DARTS~\cite{liu2018darts} searchs architecture by cells (Fig.~\ref{fig:cell}),
which is a directed acyclic graph consisting of an ordered sequence of $N$ nodes,
and it has two input nodes and a single output node \cite{zoph2017learning}.
Within a cell,
each node $x(i)$ is a latent representation
and each directed edge 
$(i, j)$ is associated with some operations 
$O(i,j)$
that transforms $x(i)$ to $x(j)$. 
Thus,
each intermediate node is computed using all of its predecessors,
i.e.,
$x^{(j)} = \sum\nolimits_{i < j} O^{(i, j)} ( x^{(i)} )$
as in Fig.~\ref{fig:cell}.
However,
such search space is discrete.
DARTS  uses softmax relaxation to make discrete choices into smooth ones (Fig.~\ref{fig:compdarts}), 
i.e.,
each $O^{(i, j)}$ is replaced by $\bar{O}^{(i, j)}$ as
\begin{align}
\bar{O}^{(i,j)}(  x^{(i)} ) 
= \nicefrac{1}{C}\sum\nolimits_{m = 1}^{|\mathcal{O}|}  
\exp(  a_{m}^{(i,j)} ) 
\mathcal{O}_m ( x^{(j)} ),
\label{eq:softmax}
\end{align}
where 
$C = \sum\nolimits_{n = 1}^{|\mathcal{O}|} \exp( a_{n}^{(i,j)} )$ is a normalization term, $\mathcal{O}_m$ denotes the $m$-th operation in search space $\mathcal{O}$.
Thus,
the choices of operations for an edge $(i,j)$ is 
replaced by a real vector $\mathbf{a}^{(i,j)} = [a_k^{(i,j)}] \in \mathbb{R}^{|\mathcal{O}|}$,
and all choices in a cell can be represented in a matrix $\mathbf{A} = [\mathbf{a}^{(i,j)}]$ 
(see Fig.~\ref{fig:alpha}). 

With such a differentiable relaxation,
the search problem in DARTS is formulated as 
\begin{align}
\min_{\mathbf{A}}  
\mathcal{L}_{\text{val}}\left(w^*, \mathbf{A} \right),
\;\text{s.t.}\;
w^* = \arg\min_{w} \mathcal{L}_{\text{train}}\left( w, \mathbf{A} \right),
\label{eq:nasdarts}
\end{align}
where $\mathcal{L}_{\text{val}}$ (resp. $\mathcal{L}_{\text{train}}$) is the loss on validation (resp. training) set,
and gradient descent is used for the optimization.
Let 
the gradient w.r.t. $\mathbf{A}$ is
\begin{align}
\nabla_{\mathbf{A}} \mathcal{L}_{\text{val}}
\left( w, \mathbf{A} \right)
= 
& \nabla_{\mathbf{A}} \mathcal{L}_{\text{val}}\left( \bar{w}(\mathbf{A}), \mathbf{A} \right)\nonumber \\  
- \varepsilon & \nabla^2_{\mathbf{A}, w} \mathcal{L}_{\text{train}}(w, \mathbf{A})
\nabla_{ \mathbf{A} } \mathcal{L}_{\text{val}}(\bar{w}, \mathbf{A}),
\label{eq:updatealpha}
\end{align}
where $\varepsilon > 0$ is the step-size and 
a second order derivative, i.e., $\nabla^2_{2, 1}(\cdot)$ is involved.
However,
the evaluation of the second order term is extremely expensive,
which requires two extra computations of gradient w.r.t. $w$ and two forward passes of $\mathbf{A}$.
Finally, a final architecture $\bar{\mathbf{A}}$ needs to be discretized from the relaxed $\mathbf{A}$
(see Alg.\ref{alg:darts}). 

\begin{algorithm}[ht]
	\caption{Differentiable architecture search (DARTS) \cite{liu2018darts}.}
	\small
	\begin{algorithmic}[1]
		\STATE Create a mixture operation $\bar{O}^{(i,j)}$ parametrized by \eqref{eq:softmax};
		\WHILE{not converged}
		
		\STATE Update $\mathbf{A}^t$ by  \eqref{eq:updatealpha};
		\\ // \textit{with second-order approximation};
		
		\STATE Update $w_t$ by $\nabla_{w_t} \mathcal{L}_{\text{train}}(w_t, \mathbf{A}^{t + 1})$
		using back-propagation;
		\\ // \textit{with all operations};
		
		\ENDWHILE
		\STATE Drive the discrete architecture $\bar{\mathbf{A}}$ from continuous $\mathbf{A}$;
		\\ // \textit{not complete};
		\RETURN final architecture $\bar{\mathbf{A}}$.
	\end{algorithmic}
	\label{alg:darts}
\end{algorithm}

Due to the differentiable relaxation in \eqref{eq:softmax},
an ensemble of operations (i.e., a supernet) are maintained and 
all operations in the search space need to be forward and backward-propagated when updating $w$;
the evaluation of the second order term in \eqref{eq:updatealpha} is very expensive known as a computation bottleneck of DARTS \cite{xie2018snas,noy2019asap}.
Besides,
the performance obtained from DARTS is also not as good as desired.
Due to the possible correlations among operations and the need of deriving a new architecture 
after the search (i.e., lack of completeness) \cite{xie2018snas}.
Finally,
the objective \eqref{eq:nasdarts} in DARTS does not 
consider model complexity,
which means DARTS cannot control the model size of the final architectures.

\section{Our Approach: NASP }

As introduced in Sec.\ref{sec:reldnas},
DARTS is a state-of-the-art NAS method, 
however, it has three significant limitations:
\begin{itemize}[leftmargin = 15px]
\item[a).] \textit{search efficiency}: the supernet resulting obtained from softmax trick in \eqref{eq:softmax} is expensive to train;

\item[b).] \textit{architecture performance}: correlations exist in operations,
which can lead to inferior architectures. 

\item[c).] \textit{model complexity}: depending on applications,
we may also want to trade accuracy for smaller models;
however, this is not allowed in DARTS.
\end{itemize}
Recall that
in earlier works of NAS
(see Tab.~\ref{tab:nasalgs}), 
e.g., NASNet \cite{baker2017designing,zoph2017neural} and GeNet \cite{xie2017genetic},
architectures are discrete when updating networks' parameters.
Such discretization naturally avoids the problem of completeness and 
correlations among operations compared with DARTS.
Thus,
\textit{can we search in a differentiable space but keep discrete architectures when updating network's parameters}?
This motivates us 
to formulate a new search objective for NAS (Sec.\ref{sec:relax}),
and propose a new algorithm for optimization (Sec.\ref{sec:proxNAS}).

\subsection{Search Objective}
\label{sec:relax}

As NAS can be seen as a black-box optimization problem \cite{yao2018taking,automl_book},
here,
we bring the wisdom of constraint optimization to deal with the NAS problem.

\subsubsection{Discrete constraint}
Specifically,
we  keep $\mathbf{A}$ being continuous,
which allows the usage of gradient descent,
but constrain the values of $\mathbf{A}$ to be discrete ones.
Thus,
we propose to use the following relaxation instead of \eqref{eq:softmax} on architectures:
\begin{align}
\!\!\!\!
\bar{O}^{(i,j)}(  x^{(i)} )  
\! = \! \sum\nolimits_{k = 1}^{|\mathcal{O}|} a^{(i,j)}_k \mathcal{O}_k(  x^{(j)} ),
\;\text{s.t.}\; 
\mathbf{a}^{(i,j)} \! \in \! \mathcal{C},
\label{eq:paraours}
\end{align}
where the constraint set is defined as
$\mathcal{C}
= \left\lbrace \mathbf{a} \,|\, \NM{\mathbf{a}}{0} = 1, \text{and}\; 0 \le a_k \le 1 \right\rbrace$.
While $\mathbf{a}^{(i,j)}$ is continuous,
the constraint $\mathcal{C}$ keeps its choices to be discrete,
and there is one operation actually activated for each edge  during training network parameter $w$ as illustrated in Fig.~\ref{fig:proxnas}.

\subsubsection{Regularization on model complexity}
Besides,
in the search space of NAS, 
different operations have distinct number of parameters. 
For example, the parameter number of "sep\_conv\_7x7" is ten times that of operation "conv\_1x1". 
Thus,
we may also want to regularize model parameters to trade-off between
accuracy and model complexity \cite{cai2018proxylessnas,xie2018snas}.

Recall that,
one column in $\mathbf{A}$ denotes one possible operation (Fig.~\ref{fig:alpha}),
and whether one operation will be selected depending on its value $\mathbf{a}^{(i,j)}$ (a row in $\mathbf{A}$).
Thus,
if we suppress the value of a specific column in $\mathbf{A}$,
its operation will be less likely to be selected in Alg.\ref{alg:proxnas},
due to the proximal step on $\mathcal{C}_1$.
These motivate us to introduce
a regularizer $\mathcal{R}(\mathbf{A})$ as
\begin{align}
\mathcal{R}(\mathbf{A}) 
=
\sum\nolimits_{k = 1}^{|\mathcal{O}|} \nicefrac{p_k}{\bar{p}} 
\NM{\dot{\mathbf{a}}_k}{2}^2, 
\label{eq:newreg}
\end{align}
where $\dot{\mathbf{a}}_k$ is the $k$th 
column in $\mathbf{A}$,
the parameter number with the $i$th operation is $p_i$,
and $\bar{p} = \sum\nolimits_{i = 1}^{|\mathcal{O}|} p_i$.

\subsubsection{Search objective}
Finally,
the NAS problem, 
with our new relaxation \eqref{eq:paraours} and regularization \eqref{eq:newreg}, 
becomes
\begin{align}
\min_{\mathbf{A}}  
\;&\;
\mathcal{F}\left(w^*, \mathbf{A} \right), 
\text{s.t.}
\begin{cases}
w^* 
\! = \! 
\underset{w}{\arg\min} 
\mathcal{L}_{\text{train}}\left( w, \mathbf{A} \right)
\\
\mathbf{a}^{(i,j)} \in \mathcal{C}
\end{cases}
\!\!\!\!,
\label{eq:proxnasreg}
\end{align}
where
$\mathcal{F} (w, \mathbf{A}) = \mathcal{L}_{\text{val}}\left(w, \mathbf{A} \right)
+ \eta \mathcal{R}(\mathbf{A})$
with $\eta \ge 0$  balancing between the complexity and the accuracy,
and a larger $\eta$ leads to smaller architectures.

\begin{remark}
Literally,
learning with a discrete constraint
has only been explored with parameters, 
e.g., deep networks compression with binary weights \cite{courbariaux2015binaryconnect},
and gradient quantization
\cite{alistarh2017qsgd},
but not in hyper-parameter or architecture optimization.
Meanwhile,
other constraints have been considered in NAS,
e.g., memory cost and latency \cite{tan2018mnasnet,cai2018proxylessnas}.
We are the first to introduce searched constraints on architecture into NAS (Tab.~\ref{tab:nasalgs}).
\end{remark}

\subsection{Search Algorithm}
\label{sec:proxNAS}

Solving the new NAS objective \eqref{eq:proxnasreg} here is not trivial.
Due to the extra constraint and regularizer,
neither simple adaptation of DARTS nor PA can be applied.
In the sequel,
we propose a new variant of PA algorithm for efficient optimization.

\subsubsection{Failure of existing algorithms}
A direct solution would be PA mentioned in Sec.\ref{sec:relprox},
then architecture  $\mathbf{A}_{t + 1}$ can be either updated by \eqref{eq:proxiter},
i.e.,
\begin{align}
\mathbf{A}_{t + 1}
= \Px{\mathcal{C}}{\mathbf{A}_t - \varepsilon \nabla_{ \bar{\mathbf{A}}_t } \mathcal{F}
	\left( w(\mathbf{A}_t), \mathbf{A}_t \right)},
\label{eq:nasp:prox1}
\end{align}
or updated by lazy proximal step \eqref{eq:proxiter:lazy}, i.e.,
\begin{align}
\bar{\mathbf{A}}_t 
& = \Px{\mathcal{C}}{ \mathbf{A}_t },
\notag
\\
\mathbf{A}_{t + 1}
& = \mathbf{A}_t - \varepsilon \nabla_{ \bar{\mathbf{A}}_t } 
\mathcal{F}( w(\bar{\mathbf{A}}_t), \bar{\mathbf{A}}_t ),
\label{eq:nasp:prox2}
\end{align}
where the gradient can be evaluated by \eqref{eq:updatealpha}
and computation of second-order approximation is still required.
Let 
$\mathcal{C}_1
= \left\lbrace \mathbf{a} \,|\, \NM{\mathbf{a}}{0} = 1 \right\rbrace$
and 
$\mathcal{C}_2 = \left\lbrace \mathbf{a} \,|\, 0 \le a_k \le 1 \right\rbrace$,
i.e., $\mathcal{C} = \mathcal{C}_1 \cap \mathcal{C}_2$.
The closed-form solution on proximal step is offered in Proposition 1 (Proofs in Appendix~A.1).

\begin{proposition} \label{pr:proxc}
$\Px{\mathcal{C}}{\mathbf{a}} 
= \Px{\mathcal{C}_1}
{ \Px{\mathcal{C}_2}{\mathbf{a}}  }$.
\end{proposition}

However,
solving \eqref{eq:proxnasreg} is not easy.
Due to the discrete nature of the constraint set,
proximal iteration \eqref{eq:nasp:prox1} is hard to obtain a good solution \cite{courbariaux2015binaryconnect}.
Besides,
while \eqref{eq:proxiter:lazy} empirically leads to better performance than \eqref{eq:proxiter}
in binary networks \cite{courbariaux2015binaryconnect,hou2016loss,Bai2018},
lazy-update \eqref{eq:nasp:prox2} will not success here neither.
The reason is that, 
as in DARTS \cite{liu2018darts},
$\mathbf{A}_t$ is naturally in range $[0, 1]$ but \eqref{eq:nasp:prox2} can not guarantee that.
This in turn will bring negative impact on the searching performance.

\subsubsection{Proposed algorithm}
Instead,
motivated by Proposition 1,
we keep $\mathbf{A}$ to be optimized as continuous variables but constrained by $\mathcal{C}_2$.
Similar box-constraints have been explored in
sparse coding and non-negative matrix factorization \cite{Lee1999LearningTP},
which help to improve the discriminative ability of learned factors.
Here,
as demonstrated in experiments,
it helps to identify better architectures.
Then, we also introduce
another discrete $\bar{\mathbf{A}}$ constrained by $\mathcal{C}_1$ derived from $\mathbf{A}$ during iterating. 
Note that, it is easy to see
$\bar{\mathbf{A}}_t \in \mathcal{C}$ is guaranteed.
The proposed procedure is described in Alg.\ref{alg:proxnas}.

\begin{table*}[ht]
	\centering
	\caption{Classification errors of NASP and state-of-the-art image classifiers on CIFAR-10.}
	\small
	\begin{tabular}{ lcc c }
		\hline
		Method                                                  &    Test Error  (\%)    & Para (M) & Time (GPU days) \\ \hline
		DenseNet-BC \cite{huang2017densely}                     &          3.46          &   25.6   & ---                    \\ \hline
		NASNet-A + cutout \cite{zoph2017learning}               &          2.65          &   3.3    & 1800                   \\
		AmoebaNet + cutout \cite{real2018regularized}           &     2.55$\pm$0.05      &   2.8    & 3150                   \\
		PNAS \cite{liu2018progressive}                          &     3.41$\pm$0.09      &   3.2    & 225                    \\
		ENAS \cite{pham2018efficient}                           &          2.89          &   4.6    & 0.5                    \\ \hline
		Random search + cutout  \cite{liu2018darts}             &     3.29$\pm$0.15      &   3.2    & 4                      \\
		DARTS (1st-order) + cutout \cite{liu2018darts}          &     3.00$\pm$0.14      &   3.3    & 1.5                    \\
		DARTS (2nd-order) + cutout                              &     2.76$\pm$0.09      &   3.3    & 4                      \\
		SNAS (large complexity) + cutout \cite{xie2018snas}     &          2.98          &   2.9    & 1.5                    \\
		SNAS (middle complexity) + cutout                       &     2.85$\pm$0.02      &   2.8    & 1.5                    \\
		SNAS (small complexity) + cutout                        &     3.10$\pm$0.04      &   2.3    & 1.5                    \\ \hline
		NASP (7 operations) + cutout                            &     2.83$\pm$0.09      &   3.3    & \textbf{0.1}           \\
		NASP (12 operations) + cutout                           & \textbf{2.44$\pm$0.04} &   7.4    & 0.2                    \\ \hline
	\end{tabular}
	\label{tab:cifar}
\end{table*}

\begin{algorithm}[ht]
\caption{NASP: Efficient Neural Architecture Search with Proximal Iterations.}
\small
\begin{algorithmic}[1]
	\STATE Create a mixture operation $\bar{O}^{(i,j)}$ parametrized by \eqref{eq:paraours};
	\WHILE{not converged}
	\STATE Get \textit{discrete} architectures: 
	$\bar{\mathbf{a}}^{(i,j)}_t = \Px{\mathcal{C}_1}{\mathbf{a}^{(i,j)}_t}$;
	
	\STATE Update $\mathbf{A}_{t + 1} = 
	\Px{\mathcal{C}_2}{\mathbf{A}_t - \nabla_{ \bar{\mathbf{A}}_t } 
		\mathcal{F}( w_t, \bar{\mathbf{A}}_t )}$; 
	\\ // \textit{no second-order approximation}

	\STATE Refine \textit{discrete} architectures: 
	$\bar{\mathbf{a}}^{(i,j)}_{t + 1} = \Px{\mathcal{C}_1}{\mathbf{a}^{(i,j)}_{t + 1}}$;
	
	\STATE Update $w_t$ by $\nabla_{w_t} \mathcal{L}_{\text{train}}(w_t, \bar{\mathbf{A}}^{t + 1})$
	using back-propagation;
	\\ // \textit{with the selected operations}
	
	\ENDWHILE
	\RETURN Searched architecture $\bar{\mathbf{A}}_t$.
\end{algorithmic}
\label{alg:proxnas}
\end{algorithm}

Compared with DARTS, NASP also alternatively updates architecture $\mathbf{A}$ (step~4)
and network parameters $w$ (step~6).
However,
note that 
$\mathbf{A}$ is discretized at step~3 and 5.
Specifically,
in step~3, 
discretized version of architectures are more stable 
than the continuous ones in DARTS,
as it is less likely for subsequent updates in $w$ to change $\bar{\mathbf{A}}$.
Thus,
we can take $w_t$ (step~4) as a constant w.r.t. $\bar{\mathbf{A}}$,
which helps us remove the second order approximation in \eqref{eq:updatealpha}
and significantly speed up architectures updates.
In step~5,
network weights need only to be propagated
with the selected operation.
This helps to reduce models' training time
and decouples operations for training networks.
Finally,
we do not need an extra step to discretize architecture from a continuous one like DARTS,
since a discrete architecture is already maintained during the search.
This helps us to reduce the gap between the search and fine-tuning, which leads to better architectures being identified.

\subsubsection{Theoretical analysis}
Finally,
unlike DARTS and PA with lazy-updates,
the convergence of the proposed NASP can be guaranteed in Theorem~\ref{thm:conv}.
The proof is in Appendix~A.2.


\begin{theorem} \label{thm:conv}
Let $\max \mathcal{F}(w, \mathbf{A}) < \infty$ and $\mathcal{F}$ be differentiable,
then the sequence $\{ \mathbf{A}^t \}$
generated by Alg.\ref{alg:proxnas} has limit points.
\end{theorem}

Note that,
previous analysis cannot be applied.
As the algorithm steps are different from 
all previous works,
i.e., \cite{hou2016loss,Bai2018},
and it is the first time that PA is introduced into NAS.
While two assumptions are made in Theorem~\ref{thm:conv},
smoothness of $\mathcal{F}$ can be satisfied using proper loss functions,
e.g., the cross-entropy in this paper,
and the second assumption can empirically hold in our experiments.

\section{Experiments}

Here,
we perform experiments with searching CNN and RNN structures.
Four datasets, i.e., CIFAR-10, 
ImageNet, PTB, WT2 will be utilized in our experiments (see Appendix~B.1).

\subsection{Architecture Search for CNN}
\label{sec:exp:cnn}

\subsubsection{Searching Cells on CIFAR-10}
Same as \cite{zoph2017neural,zoph2017learning,liu2018darts,xie2018snas,luo2018neural},
we search architectures on CIFAR-10 (\cite{krizhevsky2009learning}).
Following \cite{liu2018darts},
the convolutional cell consists of $N\!\!=\!\!7$ nodes,
and the network is obtained by stacking cells for $8$ times;
in the search process, we train a small network stacked by 8 cells with 50 epochs (see Appendix~B.2).
Two different search spaces are considered here.
The first one is the same as DARTS and contains 7 operations.
The second one is larger,
which contains 12 operations (see Appendix~B.3).
Besides, 
our search space for normal cell and reduction cell is different. 
For normal cell, the search space only consists of identity and convolutional operations;
for reduction cell, the search space only consists of identity and pooling operations.

Results compared with state-of-the-art NAS methods can be found in Tab.~\ref{tab:cifar},
the searched cells are in Appendix~C.2.
Note that ProxlessNAS \cite{cai2018proxylessnas}, 
Mnasnet \cite{tan2018mnasnet},
and Single Path One-Shot \cite{zi2018one} are not compared
as their codes are not available and they focus on NAS for mobile devices;
GeNet \cite{xie2017genetic} is not compared,
as its performance is much worse than ResNet.
Note that we remove the extra data augmentation for ASAP except cutout for a fair comparison.
We can see that
when in the same space (with 7 operations),
NASP has comparable performance with DARTS (2nd-order)
and is much better than DARTS (1st-order).
Then,
in the larger space (with 12 operations),
NASP is still much faster than DARTS,
with much lower test error than others.
Note that,
NASP on the larger space also has larger models,
as will be detailed in Sec.\ref{sec:exp:abla},
this is because NASP can find operations giving lower test error,
while others cannot.

\begin{table*}[ht]
	\centering
	\caption{Comparison with state-of-the-art language models on PTB (lower perplexity is better).}
	\small
	\begin{tabular}{ lcccc }
		\toprule
		\multirow{2}*{Architecture}           & \multicolumn{2}{c}{Perplexity (\%)} &   Params    &     Time     \\ \cline{2-3}
		~                                     &     valid     &        test         &     (M)     &  (GPU days)  \\ \hline
		NAS~\cite{zoph2017neural}             &       -       &        64.0         &     25      &    10,000    \\
		ENAS~\cite{pham2018efficient}         &     68.3      &        63.1         &     24      &     0.5      \\ \hline
		Random search~\cite{liu2018darts}     &     61.8      &        59.4         & \textbf{23} &      2       \\
		DARTS (1st-order)~\cite{liu2018darts} &     60.2      &        57.6         & \textbf{23} &     0.5      \\
		DARTS (2nd-order)                     & \textbf{59.7} &        56.4         & \textbf{23} &      1       \\ \hline
		NASP                                  &     59.9      &    \textbf{57.3}    & \textbf{23} & \textbf{0.1} \\ \bottomrule
	\end{tabular}
	\label{Results:ptb}
\end{table*}

\begin{figure}[ht]
	\centering 
	\includegraphics[width=0.45\columnwidth]{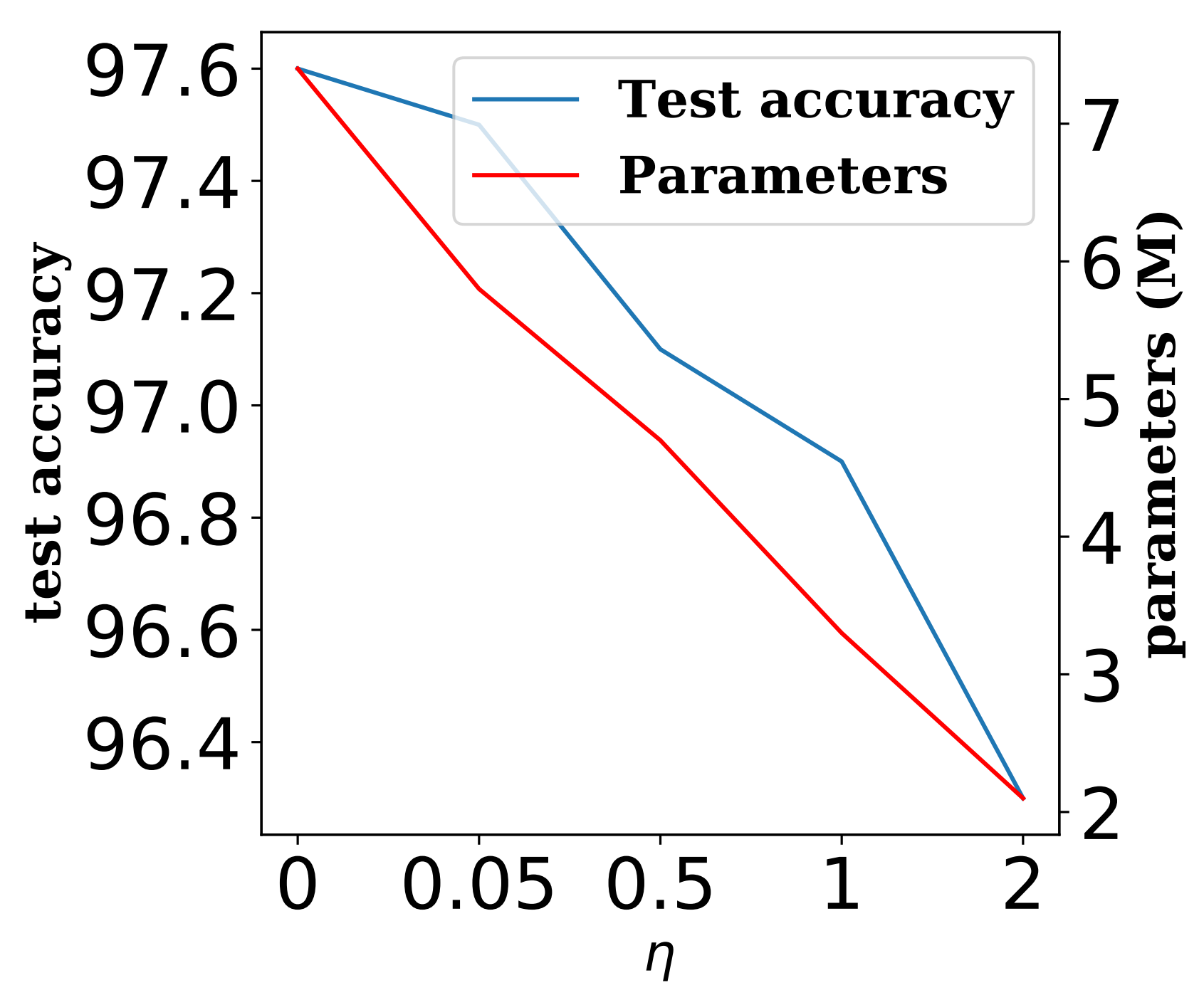}
	\vspace{-10px}
	\caption{Impact of penalty.}
    \label{Result:para_acc}
	\vspace{-10px}
\end{figure}

\textbf{Regularization on model complexity}
	In above experiments,
	we have set $\eta = 0$ for \eqref{eq:proxnasreg}.
	Here, we vary $\eta$ and the results on CIFAR-10 are demonstrated in Fig.\ref{Result:para_acc}.
	We can see that the model size gets smaller with larger $\eta$.
   
\subsubsection{Transfering to ImageNet}
In order to explore the transferability of our searched cells on ImageNet, we stack the searched cells for 14 times. 
The experiment results can be seen in Tab.~\ref{Results:imagenet}. Notably, NASP can achieve competitive test error with the state-of-the-art methods.

\begin{table}[ht]
	\centering
	\caption{Classification accuracy of NASP and state-of-the-art image classifiers on ImageNet.}
	\small
	\begin{tabular}{ L{125px} ccc}
		\hline
		\multirow{2}*{Architecture}           & \multicolumn{2}{c}{Test Error (\%)} &    Params    \\ \cline{2-3}
		~                                     & top1          &        top5         &     (M)      \\ \hline
		Inception-v1~\cite{szegedy2015going}  & 30.2          &        10.1         &     6.6      \\
		MobileNet~\cite{howard2017mobilenets} & 29.4          &        10.5         & \textbf{4.2} \\
		ShuffleNet 2~\cite{ma2018shufflenet}  & 25.1          &        10.1         &   $\sim$5    \\ \hline
		NASNet-A \cite{zoph2017learning}      & 26.8          &         8.4         &     5.3      \\
		AmoebaNet  \cite{real2018regularized} & \textbf{24.3} &    \textbf{7.6}     &     6.4      \\
		PNAS      \cite{liu2018progressive}   & 25.8          &         8.1         &     5.1      \\
		DARTS (2nd-order)                     & 26.9          &         9.0         &     4.9      \\
		SNAS (middle complexity)              & 27.3          &         9.2         &     4.3      \\ \hline
		NASP (7 operations)                   & 27.2          &         9.1         &     4.6      \\
		NASP (12 operations)                  & 26.3          &         8.6         &     9.5      \\ \hline
	\end{tabular}
	\label{Results:imagenet}
\end{table}

\subsection{Architecture Search for RNN}
\label{sec:exp:rnn}

\begin{table*}[ht]
	\centering
	\caption{Detailed comparison on computation time between DARTS and the proposed NASP on \textit{CIFAR-10}. Note that DARTS needs to search four times to obtain a good architecture \cite{liu2018darts}
    	while the performance from NASP is from one search.
    	Thus the speedup here is small than that in Table~\ref{Results:imagenet}.}
	\small
	\begin{tabular}{c | c | C{50px} | C{50px} | C{50px} | C{50px} | c || c | c}
		\hline
		           &       &                       \multicolumn{5}{c||}{computational time / epoch (in seconds)}                        &       &        \\
		  \# of    &       & \multicolumn{2}{c|}{update $\mathbf{A}$ (validation)} & \multicolumn{2}{c|}{update $w$ (training)} & total & error & params \\
		operations &       & 1st-order & $\!$2nd-order$\!$                         & forward & backward                         &       & (\%)  & (M)    \\ \hline
		    7      & DARTS & 270       & 1315                                      & 103     & 162                              & 1693  & 2.76  & 3.3    \\ \cline{2-9}
		           & NASP  & 176       & -                                         & 25      & 31                               & 343   & 2.83  & 3.3    \\ \hline
		    12     & DARTS & 489       & 2381                                      & 187     & 293                              & 3060  & 3.0   & 8.4    \\ \cline{2-9}
		           & NASP  & 303       & -                                         & 32      & 15                               & 483   & 2.5   & 7.4    \\ \hline
	\end{tabular}
	\label{tab:time}
\end{table*}

\subsubsection{Searching cells on PTB}
Following the setting of DARTS \cite{liu2018darts},
the recurrent cell consists of $N=12$ nodes; 
the first intermediate node is obtained by linearly transforming the two input nodes, adding up the results and then passing through a tanh activation function; 
then the results of the first intermediate node should be transformed by an activation function. 
The activation functions utilized are tanh, relu, sigmoid and identity. 
In the search process, we train a small network with sequence length 35 for 50 epochs. 
To evaluate the performance of searched cells on PTB, 
a single-layer recurrent network with the discovered cell is trained for at most 8000 epochs until convergence with batch size 64. Results can be seen in Tab.~\ref{Results:ptb},
and searched cells are in Appendix~C.2.
Again,
we can see DARTS's 2nd-order is much slower than 1st-order,
and NASP can be not only much faster than DARTS but also achieve 
comparable test performance with other state-of-the-art methods.

\begin{table*}[ht]
	\centering
	\caption{Classification errors of NASP and concurrent works on CIFAR-10.
		``Ops'' denotes the number of operations in the search space;
		``Nodes'' denotes the number of nodes in a cell.}
	\small
	\begin{tabular}{ lccc c c}
		\hline
		Architecture                                        &    Test Error (\%)     &   Para (M)    & Ops & Nodes & Time (GPU days) \\ \hline
		ASAP \cite{noy2019asap}                             &          3.06          &      2.6      &  7  & 4     & 0.2             \\
		ASNG \cite{akimoto2019adaptive}                     &     2.83$\pm$0.14      &      3.9      &  5  & 5     & \textbf{0.1}    \\
		BayesNAS \cite{zhou2019bayesnas}                    &     2.81$\pm$0.04      & 3.40$\pm$0.62 &  7  & 4     & 0.2             \\
		GDAS \cite{dong2019searching}                       &          2.82          & \textbf{2.5}  &  8  & 4     & 0.3             \\ \hline
		NASP (7 operations) + cutout                        &     2.83$\pm$0.09      &      3.3      &  7  & 4     & \textbf{0.1}    \\
		NASP (12 operations) + cutout                       & \textbf{2.44$\pm$0.04} &      7.4      & 12  & 4     & 0.2             \\ \hline
	\end{tabular}
	\label{tab:cifar:cc}
\end{table*}

\subsubsection{Transferring to Wiki-Text2}
Following \cite{liu2018darts},
we test the transferable ability of RNN's cell with WikiText-2 (WT2) \cite{pham2018efficient} dataset.
We train a single-layer recurrent network with the searched cells on PTB for at most 8000 epochs. 
Results can be found in Tab.~\ref{Results:wt2}.
Unlike previous case with ImageNet,
performance obtained from NAS methods are not better than human designed ones.
This is due to WT2 is harder to be transferred,
which is also observed in \cite{liu2018darts}.

\begin{table}[ht]
	\centering
	\caption{Comparison with state-of-the-art language models on WT2.
	SNAS do not provide codes on RNN and results are not reported neither.}
	\small
	\begin{tabular}{ L{120px} ccc}
		\toprule
		\multirow{2}*{Architecture}                    & \multicolumn{2}{c}{Perplexity (\%)} &   Params    \\ \cline{2-3}
		~                                              & valid         &        test         &     (M)     \\ \hline
		LSTM \cite{yang2017breaking}                   & \textbf{66.0} &    \textbf{63.3}    & \textbf{33} \\ \hline
		ENAS~\cite{pham2018efficient}                  & 72.4          &        70.4         & \textbf{33} \\
		DARTS  (2nd order)                             & 71.2          &        69.6         & \textbf{33} \\ \hline
		NASP                                           & 70.4          &        69.5         & \textbf{33} \\ \bottomrule
	\end{tabular}
	\label{Results:wt2}
\end{table}

\subsection{Ablation Study}
\label{sec:exp:abla}

\subsubsection{Comparison with DARTS}
In Sec.\ref{sec:exp:cnn},
we have shown an overall comparison between DARTS and NASP.
Here, we show detailed comparisons on updating network's parameter (i.e., $w$) and architectures (i.e., $\mathbf{A}$).
Timing results and searched performance are in Tab.~\ref{tab:time}.
First,
NASP removes much computational cost,
as no 2$nd$-order approximation of $\mathbf{A}$ and propagating $w$ with selected operations.
This clearly justifies our motivation in Sec.\ref{sec:relax}.
Second,
the discretized $\bar{\mathbf{A}}$ helps to decouple operations on updating $w$,
this helps NASP find better operations under larger search space.

\begin{figure}[ht]
	\centering
	\subfigure[\#ops = 12.]
	{\includegraphics[width=0.49\columnwidth]{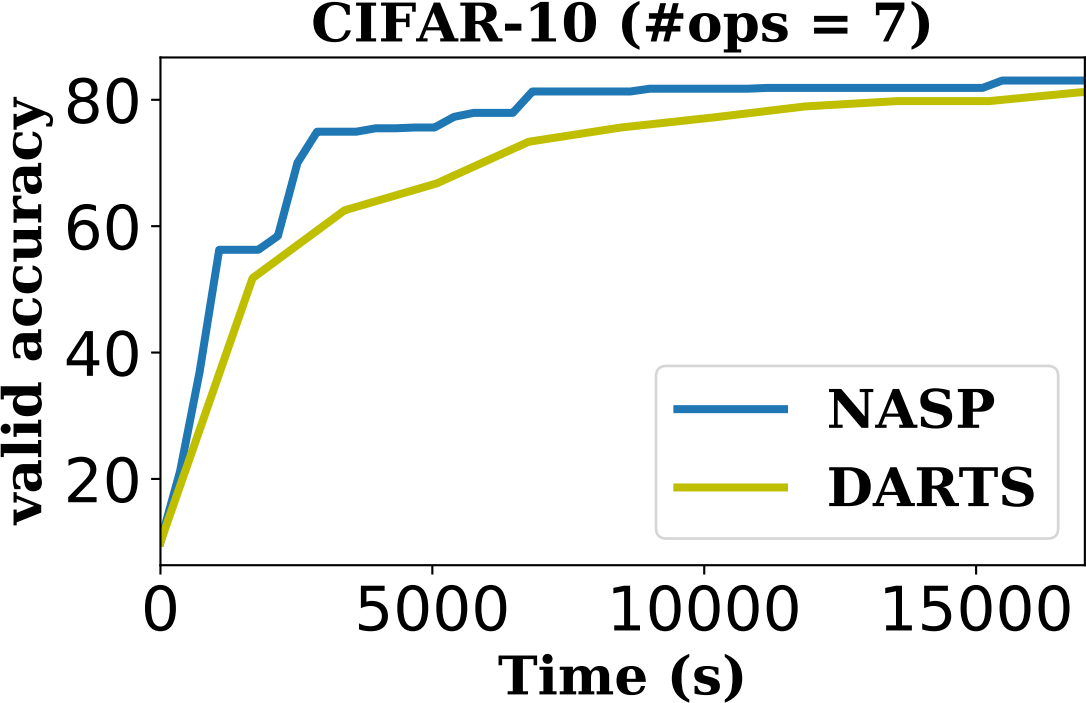}}
	\subfigure[\#ops = 7.]
	{\includegraphics[width=0.49\columnwidth]{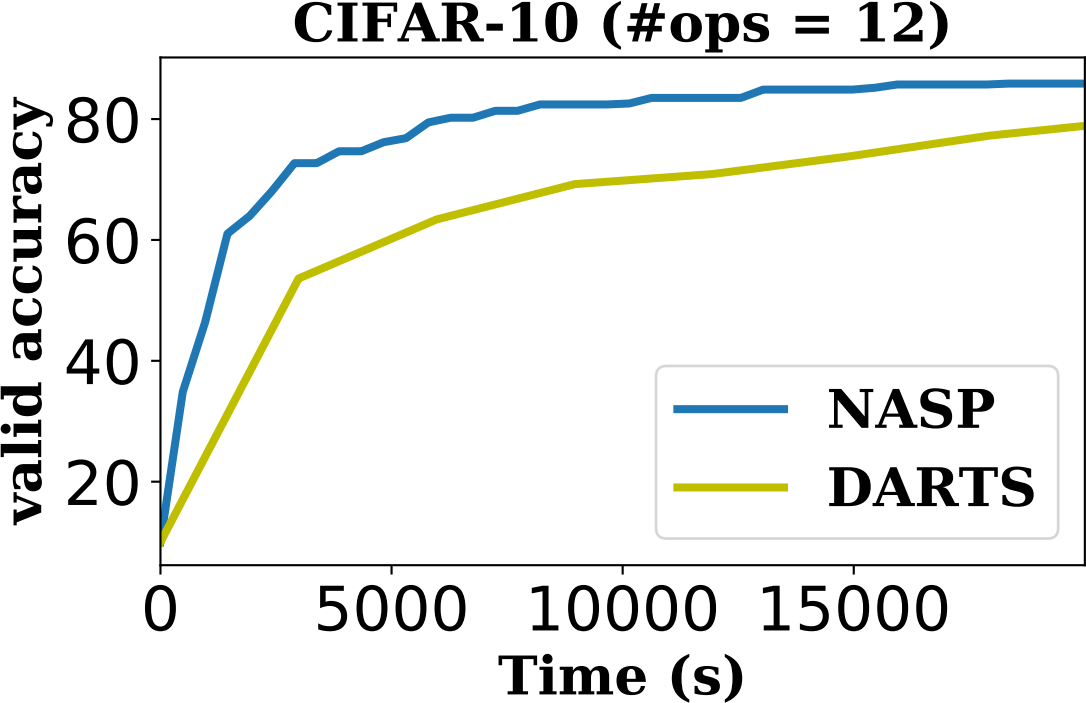}}
	
	\vspace{-10px}
	
	\caption{Comparison of NASP and DARTS on convergence.}
	\label{fig:ablation}
\end{figure}

We conduct experiments to compare the search time and validation accuracy in Fig.~\ref{fig:ablation}(a)-(b). 
We can see that in the same search time, 
our NASP obtains higher accuracy while our  NASP cost less time in the same accuracy. 
This further verifies the efficiency of NASP over DARTS. 

\begin{figure}[ht]
	\centering
	\subfigure[DARTS.]
	{\includegraphics[width=0.48\columnwidth]{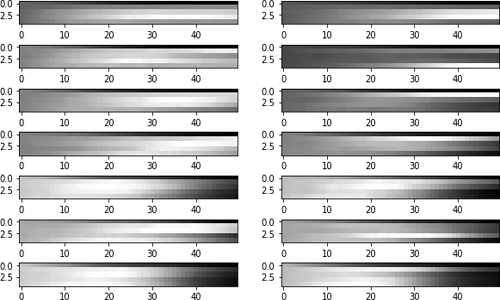}
		\label{figure:stable:darts}}
	\subfigure[NASP.]
	{\includegraphics[width=0.48\columnwidth]{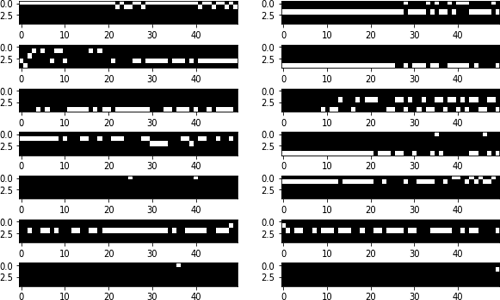}
		\label{figure:stable:nasp}}
	\vspace{-10px}
	\caption{Comparison on changes of architecture parameters between DARTS and NASP.}
	\label{figure:stable}
\end{figure}

Finally,
we illustrate why the second order approximation is a need for DARTS but not for NASP.
Recall that,
as in Sec.\ref{sec:reldnas},
as $\mathbf{A}$ continuously changes during iteration
second order approximation is to better capture $w$'s impact for $\mathbf{A}$.
Then,
in Sec.\ref{sec:proxNAS},
we argue that, 
since $\bar{\mathbf{A}}$ is discrete,
$w$'s impact will not lead to frequent changes in $\bar{\mathbf{A}}$.
This removes the needs of capturing future dynamics using the second order approximation.
We plot $\mathbf{A}$ for DARTS and $\bar{\mathbf{A}}$ for NSAP in Fig.~\ref{figure:stable}. 
In Fig.~\ref{figure:stable}, 
the x-axis represents the training epochs while the y-axis represents the operations (there are five operations selected in our figure). 
There are 14 connections between nodes, so there are 14 subfigures in both Fig.~\ref{figure:stable:darts} and \ref{figure:stable:nasp}.
Indeed,
$\bar{\mathbf{A}}$ is more stable than $\mathbf{A}$ in DARTS,
which verifies the correctness of our motivation.

\subsubsection{Comparison with standard PA}
\begin{figure}[ht]
	\centering
	\subfigure[\#ops = 12.]
	{\includegraphics[width=0.49\columnwidth]{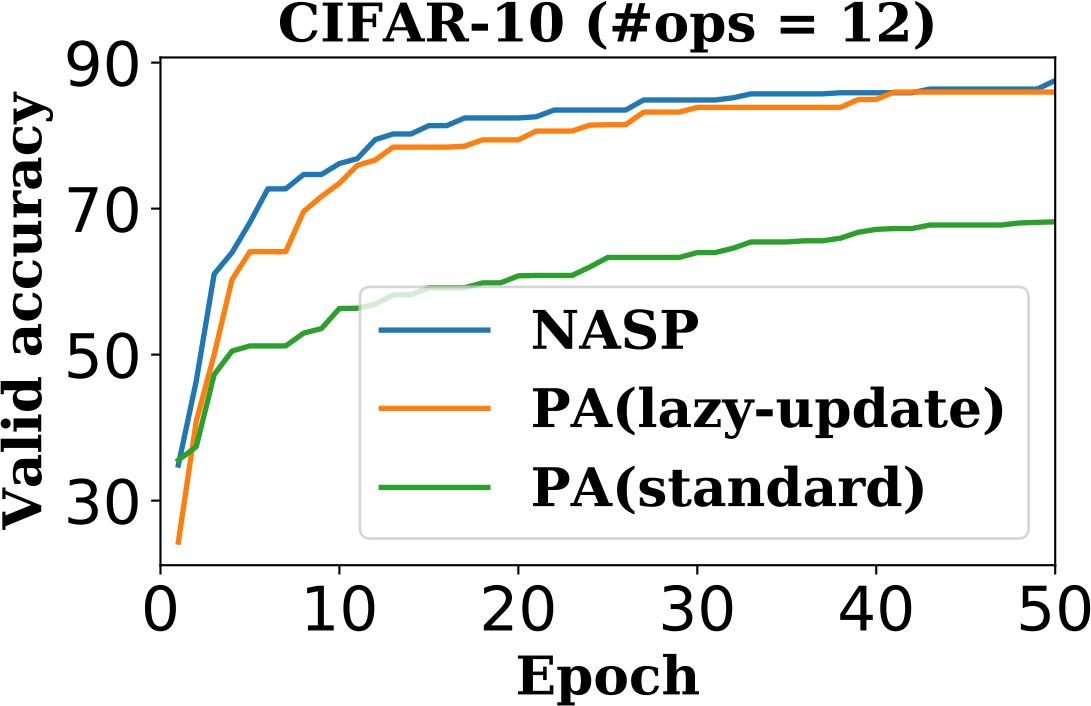}}
	\subfigure[\#ops = 7.]
	{\includegraphics[width=0.49\columnwidth]{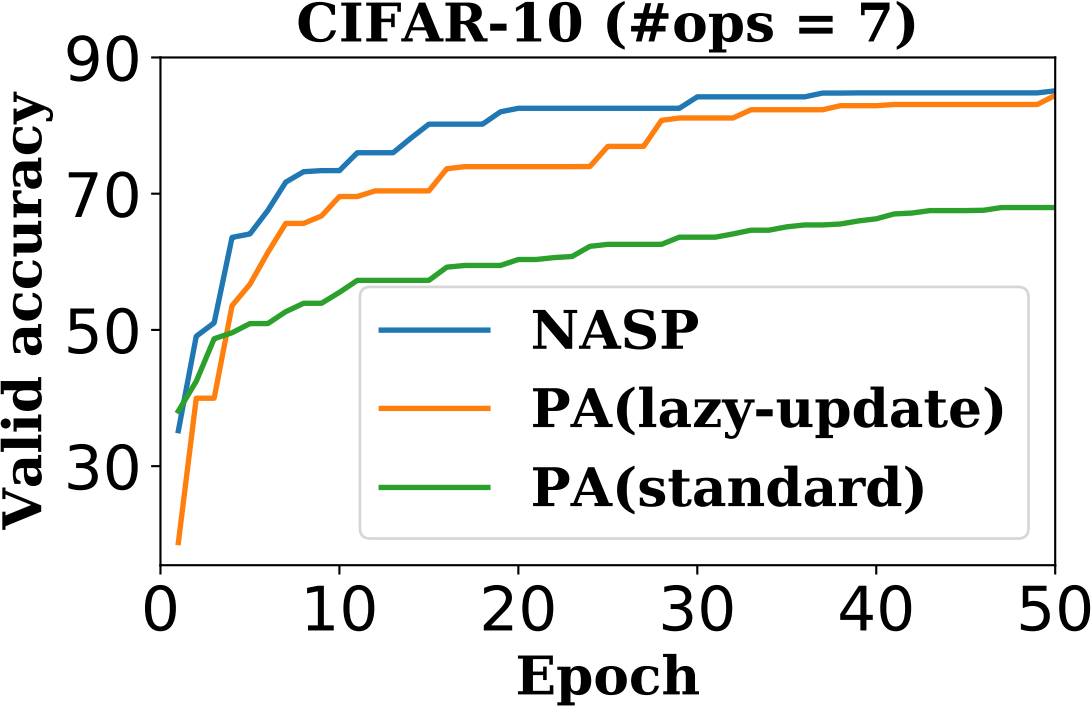}}
	\vspace{-10px}
	\caption{Comparison of NASP and direct usages of PA (i.e., simple adaptation of DARTS) on convergence.}
	\label{fig:ablation:pa}	
\end{figure}

Finally, 
we demonstrate the needs of our designs in Sec.\ref{sec:proxNAS} for NASP.
CIFAR-10 with small search space is used here.
Three algorithms are compared:
1). PA (standard), given by \eqref{eq:nasp:prox1};
2). PA (lazy-update), given by \eqref{eq:nasp:prox2};
and 3) NASP.
Results are in Fig.~\ref{fig:ablation:pa}(a) and Fig.~\ref{fig:ablation:pa}(b).
First,
good performance cannot be obtained from a direct proximal step,
which is due to the discrete constraint.
Same observation is also previous made for binary networks \cite{courbariaux2015binaryconnect}. 
Second,
PA(lazy-update) is much better than PA(standard)
but still worse than NASP.
This verifies the needs to keep elements of the matrix $\mathbf{A}$ in $[0, 1]$,
as it can encourage better operations.

\subsection{Comparison with Concurrent Works}
\label{sec:concurrent}

When we conducted our work, there were some concurrent works.
ASAP~\cite{noy2019asap} and BayesNAS~\cite{zhou2019bayesnas}
take NAS as a network pruning problem,
they remove operations that are not promising during the search.
ASNG~\cite{akimoto2019adaptive} 
and GDAS~\cite{dong2019searching} both take stochastic relaxation
to the search space, 
the difference is that ASNG uses natural gradient descent \cite{amari1998natural} for optimization
while GDAS use Gumbel-Max trick \cite{jang2016categorical} with gradient descent.
We compare NASP with them in Tab.~\ref{tab:cifar:cc} and \ref{tab:con:rnn}. 
Note that ASAP,
ASNG and BayesNAS cannot be used for searching RNN's architectures.
We can see NASP is more efficient than these works and
offer better performance on the CNN task.
Besides,
NASP can also be applied with RNN.

%
%
%
%

\begin{table}[ht]
	\centering
	\caption{Comparison of NASP with concurrent works on PTB.
		ASAP, ASNG and BayesNAS are not compared
		as they cannot be used for searching RNN.}
	\small
	\begin{tabular}{ C{40px} C{32px} C{32px} c c }
		\toprule
		\multirow{2}*{Method} & \multicolumn{2}{c}{Perplexity (\%)} & Params & Search Cost  \\ \cline{2-3}
		~                     &     valid     &        test         &  (M)   &  (GPU days)  \\ \hline
		GDAS                  & \textbf{59.8} &        57.5         &   23   &     0.4      \\ \hline
		NASP                  &     59.9      &    \textbf{57.3}    &   23   & \textbf{0.1} \\ \bottomrule
	\end{tabular}
	\label{tab:con:rnn}
\end{table}

\section{Conclusion}

We introduce NASP, a fast and differentiable neural architecture search method via proximal iterations. 
Compared with DARTS, NASP is more efficient and performs better. The key contribution of NASP is the proximal iterations in search process. 
This approach makes only one operation updated, which saves much time and makes it possible to utilize a larger search space. 
Besides, our NASP eliminates the correlation among operations. 
Experiments demonstrate that our NASP is faster and obtain better performance than baselines. 

\section*{Acknowledgments}

Q. Yao would like to thank Xiangning Chen, 
Yongqi Zhang,
and Huan Zhao for their helpful feedback.
Z. Zhu is supported in part by National Natural Science Foundation of China (No.61806009), Beijing Natural Science Foundation(No. 4184090) and Beijing Academy of Artificial Intelligence (BAAI).

{
\cleardoublepage
\bibliographystyle{aaai}
\bibliography{bib}

\begin{thebibliography}{}

\bibitem[\protect\citeauthoryear{Akimoto \bgroup et al\mbox.\egroup
  }{2019}]{akimoto2019adaptive}
Akimoto, Y.; Shirakawa, S.; Yoshinari, N.; Uchida, K.; Saito, S.; and Nishida,
  K.
\newblock 2019.
\newblock Adaptive stochastic natural gradient method for one-shot neural
  architecture search.
\newblock In {\em ICML},  171--180.

\bibitem[\protect\citeauthoryear{Alistarh \bgroup et al\mbox.\egroup
  }{2017}]{alistarh2017qsgd}
Alistarh, D.; Grubic, D.; Li, J.; Tomioka, R.; and Vojnovic, M.
\newblock 2017.
\newblock {QSGD}: Communication-efficient sgd via gradient quantization and
  encoding.
\newblock In {\em NeurIPS},  1709--1720.

\bibitem[\protect\citeauthoryear{Amari}{1998}]{amari1998natural}
Amari, S.
\newblock 1998.
\newblock Natural gradient works efficiently in learning.
\newblock {\em Neural Computation} 10(2):251--276.

\bibitem[\protect\citeauthoryear{Bai, Wang, and Liberty}{2018}]{Bai2018}
Bai, Y.; Wang, Y.-X.; and Liberty, E.
\newblock 2018.
\newblock Proxquant: Quantized neural networks via proximal operators.
\newblock In {\em ICLR}.

\bibitem[\protect\citeauthoryear{Baker \bgroup et al\mbox.\egroup
  }{2017}]{baker2017designing}
Baker, B.; Gupta, O.; Naik, N.; and Raskar, R.
\newblock 2017.
\newblock Designing neural network architectures using reinforcement learning.
\newblock In {\em ICLR}.

\bibitem[\protect\citeauthoryear{Cai, Zhu, and Han}{2019}]{cai2018proxylessnas}
Cai, H.; Zhu, L.; and Han, S.
\newblock 2019.
\newblock {ProxylessNAS}: Direct neural architecture search on target task and
  hardware.
\newblock In {\em ICLR}.

\bibitem[\protect\citeauthoryear{Courbariaux, Bengio, and
  David}{2015}]{courbariaux2015binaryconnect}
Courbariaux, M.; Bengio, Y.; and David, J.-P.
\newblock 2015.
\newblock Binaryconnect: Training deep neural networks with binary weights
  during propagations.
\newblock In {\em NeurIPS},  3123--3131.

\bibitem[\protect\citeauthoryear{Devries and
  Taylor}{2017}]{devries2017improved}
Devries, T., and Taylor, G.
\newblock 2017.
\newblock Improved regularization of convolutional neural networks with cutout.
\newblock Technical report, arXiv:1708.04552.

\bibitem[\protect\citeauthoryear{Dong and Yang}{2019}]{dong2019searching}
Dong, X., and Yang, Y.
\newblock 2019.
\newblock Searching for a robust neural architecture in four {GPU} hours.
\newblock In {\em CVPR},  1761--1770.

\bibitem[\protect\citeauthoryear{Guo \bgroup et al\mbox.\egroup
  }{2019}]{zi2018one}
Guo, Z.; Zhang, X.; Mu, H.; Heng, W.; Liu, Z.; Wei, Y.; and Sun, J.
\newblock 2019.
\newblock Single path one-shot neural architecture search with uniform
  sampling.
\newblock Technical report, Arvix.

\bibitem[\protect\citeauthoryear{He \bgroup et al\mbox.\egroup }{2016}]{He2016}
He, K.; Zhang, X.; Ren, S.; and Sun, J.
\newblock 2016.
\newblock Deep residual learning for image recognition.
\newblock In {\em CVPR},  770--778.

\bibitem[\protect\citeauthoryear{Hou, Yao, and Kwok}{2017}]{hou2016loss}
Hou, L.; Yao, Q.; and Kwok, J.
\newblock 2017.
\newblock Loss-aware binarization of deep networks.
\newblock In {\em ICLR}.

\bibitem[\protect\citeauthoryear{Howard \bgroup et al\mbox.\egroup
  }{2017}]{howard2017mobilenets}
Howard, A.; Zhu, M.; Chen, B.; Kalenichenko, D.; Wang, W.; Weyand, T.;
  Andreetto, M.; and Adam, H.
\newblock 2017.
\newblock Mobilenets: Efficient convolutional neural networks for mobile vision
  applications.
\newblock {\em CVPR}.

\bibitem[\protect\citeauthoryear{Huang \bgroup et al\mbox.\egroup
  }{2017}]{huang2017densely}
Huang, G.; Liu, Z.; Van Der~Maaten, L.; and Weinberger, K.
\newblock 2017.
\newblock Densely connected convolutional networks.
\newblock In {\em CVPR},  4700--4708.

\bibitem[\protect\citeauthoryear{Hutter, Kotthoff, and
  Vanschoren}{2018}]{automl_book}
Hutter, F.; Kotthoff, L.; and Vanschoren, J., eds.
\newblock 2018.
\newblock {\em Automated Machine Learning: Methods, Systems, Challenges}.
\newblock Springer.

\bibitem[\protect\citeauthoryear{Jang, Gu, and
  Poole}{2016}]{jang2016categorical}
Jang, E.; Gu, S.; and Poole, B.
\newblock 2016.
\newblock Categorical reparameterization with gumbel-softmax.
\newblock In {\em ICLR}.

\bibitem[\protect\citeauthoryear{Krizhevsky}{2009}]{krizhevsky2009learning}
Krizhevsky, A.
\newblock 2009.
\newblock Learning multiple layers of features from tiny images.
\newblock Technical report, Citeseer.

\bibitem[\protect\citeauthoryear{Le and Yang}{2015}]{Tiny}
Le, Y., and Yang, X.
\newblock 2015.
\newblock Tiny imagenet visual recognition challenge.
\newblock {\em CS 231N}.

\bibitem[\protect\citeauthoryear{Lee and Seung}{1999}]{Lee1999LearningTP}
Lee, D., and Seung, S.
\newblock 1999.
\newblock Learning the parts of objects by non-negative matrix factorization.
\newblock {\em Nature} 401:788--791.

\bibitem[\protect\citeauthoryear{Liu \bgroup et al\mbox.\egroup
  }{2018}]{liu2018progressive}
Liu, C.; Zoph, B.; Shlens, J.; Hua, W.; Li, L.; Li, F.-F.; Yuille, A.; Huang,
  J.; and Murphy, K.
\newblock 2018.
\newblock Progressive neural architecture search.
\newblock In {\em ECCV}.

\bibitem[\protect\citeauthoryear{Liu, Simonyan, and Yang}{2019}]{liu2018darts}
Liu, H.; Simonyan, K.; and Yang, Y.
\newblock 2019.
\newblock {DARTS}: Differentiable architecture search.
\newblock In {\em ICLR}.

\bibitem[\protect\citeauthoryear{Luo \bgroup et al\mbox.\egroup
  }{2018}]{luo2018neural}
Luo, R.; Tian, F.; Qin, T.; Chen, E.; and Liu, T.-Y.
\newblock 2018.
\newblock Neural architecture optimization.
\newblock In {\em NeurIPS}.

\bibitem[\protect\citeauthoryear{Ma \bgroup et al\mbox.\egroup
  }{2018}]{ma2018shufflenet}
Ma, N.; Zhang, X.; Zheng, H.; and Sun, J.
\newblock 2018.
\newblock {ShuffleNet V2}: Practical guidelines for efficient {CNN}
  architecture design.
\newblock {\em ECCV}  122--138.

\bibitem[\protect\citeauthoryear{Noy \bgroup et al\mbox.\egroup
  }{2019}]{noy2019asap}
Noy, A.; Nayman, N.; Ridnik, T.; Zamir, N.; Doveh, S.; Friedman, I.; Giryes,
  R.; and Zelnik-Manor, L.
\newblock 2019.
\newblock {ASAP}: Architecture search, anneal and prune.
\newblock Technical report, arXiv preprint arXiv:1904.04123.

\bibitem[\protect\citeauthoryear{Parikh and Boyd}{2013}]{parikh2013proximal}
Parikh, N., and Boyd, S.
\newblock 2013.
\newblock Proximal algorithms.
\newblock {\em Foundations and Trends in Optimization} 1(3):123--231.

\bibitem[\protect\citeauthoryear{Pham \bgroup et al\mbox.\egroup
  }{2018}]{pham2018efficient}
Pham, H.; Guan, M.; Zoph, B.; Le, Q.; and Dean, J.
\newblock 2018.
\newblock Efficient neural architecture search via parameter sharing.
\newblock Technical report, arXiv preprint.

\bibitem[\protect\citeauthoryear{Real \bgroup et al\mbox.\egroup
  }{2018}]{real2018regularized}
Real, E.; Aggarwal, A.; Huang, T.; and Le, Q.
\newblock 2018.
\newblock Regularized evolution for image classifier architecture search.
\newblock {\em arXiv}.

\bibitem[\protect\citeauthoryear{Sutton and
  Barto}{1998}]{sutton1998reinforcement}
Sutton, R., and Barto, A.
\newblock 1998.
\newblock {\em Reinforcement learning: An introduction}.
\newblock MIT press.

\bibitem[\protect\citeauthoryear{Szegedy \bgroup et al\mbox.\egroup
  }{2015}]{szegedy2015going}
Szegedy, C.; Liu, W.; Jia, Y.; Sermanet, P.; Reed, S.~E.; Anguelov, D.; Erhan,
  D.; Vanhoucke, V.; and Rabinovich, A.
\newblock 2015.
\newblock Going deeper with convolutions.
\newblock {\em CVPR}  1--9.

\bibitem[\protect\citeauthoryear{Tan \bgroup et al\mbox.\egroup
  }{2018}]{tan2018mnasnet}
Tan, M.; Chen, B.; Pang, R.; Vasudevan, V.; and Le, Q.
\newblock 2018.
\newblock Mnasnet: Platform-aware neural architecture search for mobile.
\newblock Technical report, arXiv.

\bibitem[\protect\citeauthoryear{Xiao}{2010}]{xiao2010dual}
Xiao, L.
\newblock 2010.
\newblock Dual averaging methods for regularized stochastic learning and online
  optimization.
\newblock {\em JMLR} 11(Oct):2543--2596.

\bibitem[\protect\citeauthoryear{Xie and Yuille}{2017}]{xie2017genetic}
Xie, L., and Yuille, A.
\newblock 2017.
\newblock Genetic {CNN}.
\newblock In {\em ICCV}.

\bibitem[\protect\citeauthoryear{Xie \bgroup et al\mbox.\egroup
  }{2019}]{xie2018snas}
Xie, S.; Zheng, H.; Liu, C.; and Lin, L.
\newblock 2019.
\newblock {SNAS}: stochastic neural architecture search.
\newblock In {\em ICLR}.

\bibitem[\protect\citeauthoryear{Yang \bgroup et al\mbox.\egroup
  }{2018}]{yang2017breaking}
Yang, Z.; Dai, Z.; Salakhutdinov, R.; and Cohen, W.
\newblock 2018.
\newblock Breaking the softmax bottleneck: A high-rank rnn language model.
\newblock In {\em ICLR}.

\bibitem[\protect\citeauthoryear{Yao \bgroup et al\mbox.\egroup
  }{2017}]{yao2017efficient}
Yao, Q.; Kwok, J.; Gao, F.; Chen, W.; and Liu, T.-Y.
\newblock 2017.
\newblock Efficient inexact proximal gradient algorithm for nonconvex problems.
\newblock In {\em IJCAI},  3308--3314.
\newblock AAAI Press.

\bibitem[\protect\citeauthoryear{Yao \bgroup et al\mbox.\egroup
  }{2018}]{yao2018taking}
Yao, Q.; Wang, M.; Chen, Y.; Dai, W.; Hu, Y.; Li, Y.; Tu, W.-W.; Yang, Q.; and
  Yu, Y.
\newblock 2018.
\newblock Taking human out of learning applications: A survey on automated
  machine learning.
\newblock Technical report, arXiv preprint.

\bibitem[\protect\citeauthoryear{Zhong \bgroup et al\mbox.\egroup
  }{2018}]{zhong2018practical}
Zhong, Z.; Yan, J.; Wu, W.; Shao, J.; and Liu, C.-L.
\newblock 2018.
\newblock Practical block-wise neural network architecture generation.
\newblock In {\em CVPR}.

\bibitem[\protect\citeauthoryear{Zhou \bgroup et al\mbox.\egroup
  }{2019}]{zhou2019bayesnas}
Zhou, H.; Yang, M.; Wang, J.; and Pan, W.
\newblock 2019.
\newblock {BayesNAS}: A bayesian approach for neural architecture search.
\newblock In {\em ICML},  7603--7613.

\bibitem[\protect\citeauthoryear{Zoph and Le}{2017}]{zoph2017neural}
Zoph, B., and Le, Q.
\newblock 2017.
\newblock Neural architecture search with reinforcement learning.
\newblock In {\em ICLR}.

\bibitem[\protect\citeauthoryear{Zoph \bgroup et al\mbox.\egroup
  }{2017}]{zoph2017learning}
Zoph, B.; Vasudevan, V.; Shlens, J.; and Le, Q.
\newblock 2017.
\newblock Learning transferable architectures for scalable image recognition.
\newblock In {\em CVPR}.

\end{thebibliography}
}


\appendix

\section{Proofs}

\subsection{Proposition 1}
\label{app:proposition1}
\begin{proof}
	Recall that $\mathcal{C} = \mathcal{C}_1 \cap \mathcal{C}_2$ where
	$\mathcal{C}_1
	= \left\lbrace \mathbf{a} \,|\, \NM{\mathbf{a}}{0} = 1 \right\rbrace$
	and
	$\mathcal{C}_2 = \left\lbrace \mathbf{a} \,|\, 0 \le a_k \le 1 \right\rbrace $,
	and the proximal step is given by
	\begin{align}
	\Px{\mathcal{C}}{\mathbf{a}} 
	& = \mathbf{b}^* =  \arg\min_{\mathbf{b}}\frac{1}{2}
	\NM{\mathbf{a} - \mathbf{b}}{2}^2,
	\label{app:prox}
	\\
	& \;\text{s.t.}\; \mathbf{a} \in \mathcal{C}_1 \cap \mathcal{C}_2.
	\notag
	\end{align}
	Constrain $\mathcal{C}_1$ is means $\mathbf{b}^*$ can be represented as 
	$c \, \mathbf{e}_i$ where $c$ is a parameter to be determined and 
	$\mathbf{e}_i$ is a one-hot vector with the $i$th element being $1$ and all others are zeros;
	moreover,
	constrain $\mathcal{C}_2$ means $c$ must be in range $[0, 1]$.
	Let $\mathbf{a}$ be a $k$-dimensional vector,
	then \eqref{app:prox} can be decomposed as $k$ separable problems,
	i.e.,
	\begin{align}
	c^*_i
	= \arg\min_{c}\frac{1}{2}
	\NM{\mathbf{a} - c \, \mathbf{e}_i}{2}^2,
	\text{\;for\;}
	i = 1, \dots, k.
	\end{align}
	where
	\begin{align*}
	c^*_i 
	= \begin{cases}
	0 & \text{if}\; a_i < 0
	\\
	a_i & \text{if}\; 0 \le a_i \le 1
	\\
	1 & \text{otherwise}
	\end{cases}.
	\end{align*}
	Let
	$v_i = \frac{1}{2}\left( a_i - c_i^* \right)^2$ and $\mathbf{v} = [v_1, \dots, v_k]$.
	These contain all $k$ possible solutions for \eqref{app:prox}.
	Thus, 
	in order to achieve the minimum, 
	we need to pick up $i^* = \arg\max_{i} v_i$,
	and $\mathbf{b}^* = c^*_{i^*} \mathbf{e}_{i^*}$.
	More compactly,
	we can express it as
	\begin{align*}
	\Px{\mathcal{C}}{\mathbf{a}} 
	= \Px{\mathcal{C}_1}
	{ \Px{\mathcal{C}_2}{\mathbf{a}}  }.
	\end{align*}
\end{proof}

\subsection{Theorem~\ref{thm:conv}}

\begin{proof}
Note that by the definition of loss functions,
\begin{itemize}
\item $\mathcal{F}$ is the loss on the validation set,
thus $\mathcal{F}$ is bounded from below;

\item $\mathcal{F}$ is continuous on $\mathbf{A}$.
\end{itemize} 
Since 
$\mathbf{A}_t \in \mathcal{C}_2$ and $\max \mathcal{F}(w_t, \mathbf{A}_t) < \infty$,
thus $\{ \mathbf{A}_t \}$ is constrained within a compact sublevel set. 
Finally,
the Theorem comes from the fact that 
any infinite sequences on a compact sub-level set 
must have limit points.
\end{proof}

\begin{table*}[ht]
	\centering
	\caption{Classification accuracy of NASP and state-of-the-art image classifiers on Tiny ImageNet.}
	\small
	\begin{tabular}{ lcccc }
		\hline
		\multirow{2}*{Method}                      & \multicolumn{2}{c}{Test Accuracy (\%)} &    Params    & Search Cost  \\ \cline{2-3}
		~                                                &      top1      &         top5          &     (M)      &  (GPU days)  \\ \hline
		ResNet18 \cite{He2016}                           &     52.67      &         76.77         &     11.7     &      -       \\ \hline
		NASNet-A \cite{zoph2017learning}                 & \textbf{58.99} &    \textbf{77.85}     &     4.8      &     1800     \\
		AmoebaNet-A \cite{real2018regularized}           &     57.16      &         77.62         &     4.2      &     3150     \\
		ENAS \cite{pham2018efficient}                    &     57.81      &         77.28         &     4.6      &     0.5      \\
		DARTS   \cite{liu2018darts}                      &     57.42      &         76.83         &     3.9      &      4       \\
		SNAS   \cite{xie2018snas}                        &     57.81      &         76.93         &     3.3      &     1.5      \\ \hline
		NASP                                             &     58.12      &         77.62         &     4.0      & \textbf{0.1} \\
		NASP (more ops)                                  &     58.32      &         77.54         &     8.9      &     0.2      \\ \hline
	\end{tabular}
	\label{Results:tinyimagenet}
\end{table*}

%
%

\begin{figure*}[ht]
\centering
\includegraphics[width = 0.75\textwidth]{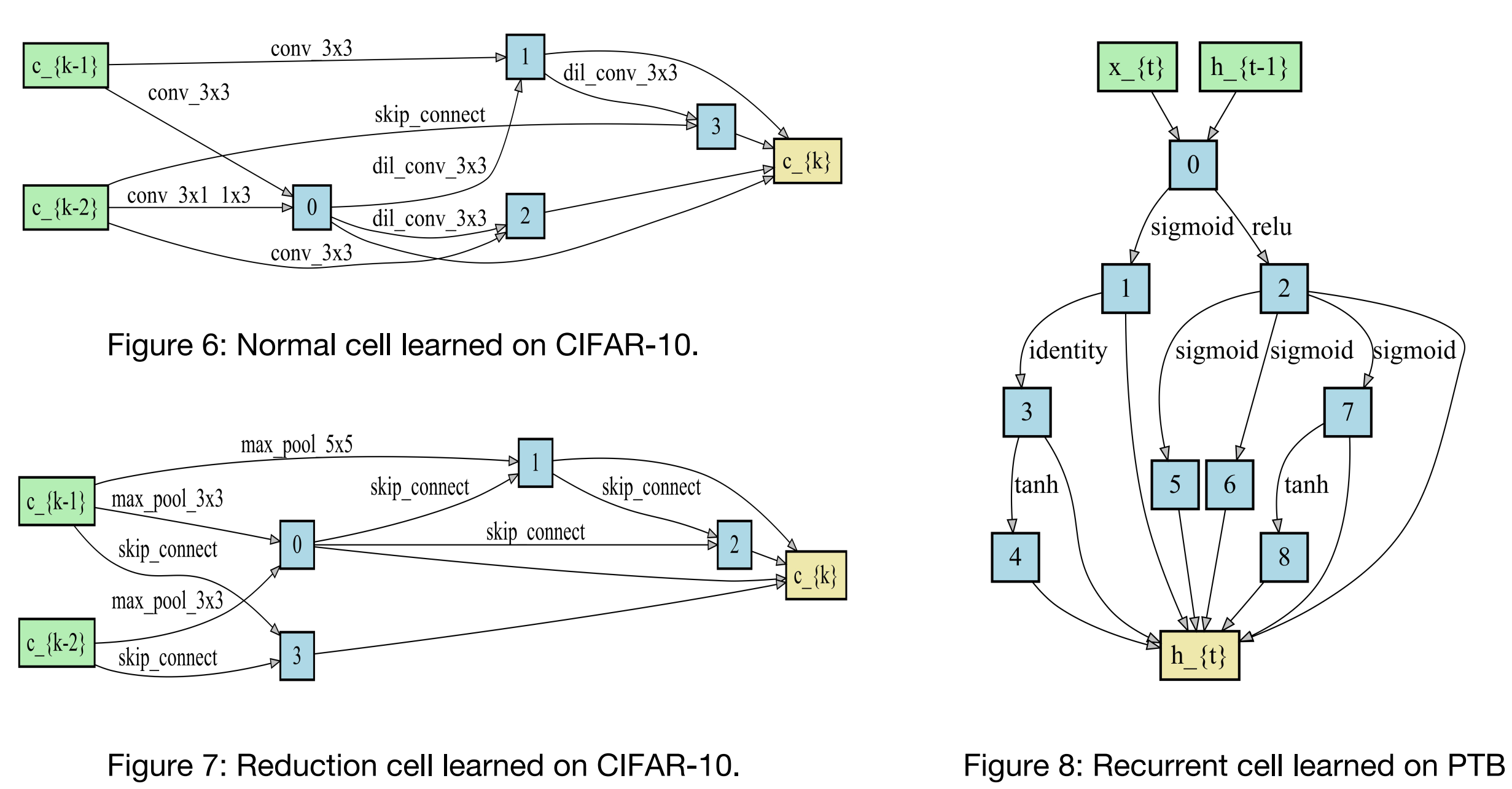}
\end{figure*}

\section{Experiment Details}

\subsection{Datasets}
\label{app:datasets}

\subsubsection{CIFAR-10}
CIFAR-10~\cite{krizhevsky2009learning}\footnote{\url{http://www.cs.toronto.edu/~kriz/cifar.html}}
is a basic dataset for image classification, which consists of 50,000 training images and 10,000 testing images.
Half of the CIFAR-10 training images will be utilized as the validation set. 
Data augmentation like cutout \cite{devries2017improved} and
HorizontalFlip will be utilized in our experiments. After training, we will test the model on test dataset and report accuracy in our experiments.

\subsubsection{PTB}
PTB\footnote{\url{http://www.fit.vutbr.cz/~imikolov/rnnlm/simple-examples.tgz}}
is an English corpus used for probabilistic language modeling, which consists of approximately 7 million words of part-of-speech tagged text, 
3 million words of skeletally parsed text, over 2 million words of text parsed for predicate-argument structure, 
and 1.6 million words of transcribed spoken text annotated for speech dis-fluencies. 
We will choose the model with the best performance on validation dataset and test it on test dataset.

\subsubsection{WT2}
Compared to the preprocessed version of Penn Treebank (PTB), 
\footnote{\url{https://s3.amazonaws.com/research.metamind.io/wikitext/wikitext-2-v1.zip}}WikiText-2 (WT2) 
is over 2 times larger. WT2 features a far larger vocabulary and retains the original case, punctuation and numbers 
- all of which are removed in PTB.As it is composed of full articles, the dataset is well suited for models that can take advantage of long term dependencies. 
We will choose the model with the best performance on validation dataset and test it on test dataset.

%
%
%
%
%
%
%
%
%
%

\subsection{Training details}
\label{app:traincnn}
For training CIFAR-10, the convolutional cell consists of $N\!\!=\!\!7$ nodes,
and the network is obtained by stacking cells for $8$ times;
in the search process, we train a small network stacked by 8 cells with 50 epochs.
SGD is utilized to optimize the network's weights, 
and Adam is utilized for the parameters of network architecture. 
To evaluate the performance of searched cells, 
the searched cells are stacked for 20 times; 
the network will be fine-tuned for 600 epochs with batch size 96. 
Additional enhancements like path dropout
(of probability 0.2) 
and auxiliary towers (with weight 0.4) are also used. We have run our experiments for three times and report the mean.

\subsection{Search Space}
\label{app:searchspace}

NASP's search space:
identity, 1x3 then 3x1 convolution, 3x3 dilated convolution, 3x3 average pooling, 3x3 max pooling, 5x5 max pooling, 
7x7 max pooling, 1x1 convolution, 3x3 convolution, 3x3 depthwise-separable conv, 5x5 depthwise-seperable conv, 7x7 depthwise-separable conv.

\section{More Experiments}

\subsection{Transferring to Tiny ImageNet}
\label{sec:tiny}

\subsubsection{Dataset}

Tiny ImageNet~\cite{Tiny}\footnote{\url{http://tiny-imagenet.herokuapp.com/}}
contains a training set of 100,000 images, a testing set of 10,000 images. 
These images are sourced from 200 different classes of objects from ImageNet.
Note that due to small number of training images for each class and low-resolution for images, Tiny ImageNet is harder to be trained than the original ImageNet. 
Data augmentation like RandomRotation and RandomHorizontalFlip are utilized. After training, we will test the model on test dataset and report accuracy in our experiments.

\subsubsection{Results}
The architecture transferability is important for cells to transfer to other datasets \cite{zoph2017learning}. 
To explore the transferability of our searched cells, we stack searched cells for 14 times  on Tiny ImageNet, 
and train the network for 250 epochs. 
Results are in Tab.~\ref{Results:tinyimagenet}. 
We can see NASP exhibits good transferablity,
and its performance is also better than other methods except NASNet-A. But our NASP is much faster than NASNet-A.

\subsection{Searched Architectures}
\label{app:seararch}

Architectures identified on CIFAR-10 are shown in Fig.6 and 7,
on PTB is shown in Fig.8.

\end{document}